\documentclass[10pt,twocolumn,letterpaper]{article}

\usepackage{cvpr}              
\usepackage{times}  
\usepackage{helvet}  
\usepackage{courier}  
\usepackage[hyphens]{url}  
\usepackage{graphicx} 
\usepackage{natbib}  
\usepackage{caption} 

\usepackage{multirow}
\usepackage{booktabs}
\usepackage{caption}
\usepackage{graphicx}
\usepackage{amsmath, bm}
\usepackage{lipsum}
\usepackage{microtype}
\usepackage{mathrsfs}
\usepackage{fancyhdr}
\usepackage{enumitem}

\usepackage{amssymb}
\usepackage{booktabs}
\usepackage[table]{xcolor}
\usepackage{makecell} 
\usepackage{booktabs}  
\usepackage{xcolor}    
\usepackage{booktabs,tabularx}
\usepackage{amsmath}     
\usepackage{amsfonts}    
\usepackage{algorithm}   
\usepackage{algpseudocode} 
\usepackage{amsmath}
\usepackage{marvosym}
\usepackage{dsfont} 
\definecolor{cvprblue}{rgb}{0.21,0.49,0.74}
\usepackage[pagebackref,breaklinks,colorlinks,allcolors=cvprblue]{hyperref}

\def\paperID{30130} 
\def\confName{CVPR}
\def\confYear{2026}

\title{Adaptive Learned Image Compression with Graph Neural Networks}

\author{%
  Yunuo Chen$^{1}$
  ~~Bing He$^{1}$
  ~~Zezheng Lyu$^{2}$
  ~~Hongwei Hu$^{3}$
  ~~Qunshan Gu$^{3}$ 
  ~~Yuan Tian$^{4}$
  ~~Guo Lu$^{1}$  \footnotemark[1] 
  \\
  $^{1}$Shanghai Jiao Tong University ~~ \quad
  $^{2}$Massachusetts Institute of Technology\\
  $^{3}$Alibaba Group ~~\quad
  $^{4}$Shanghai AI Laboratory
}

\begin{document}
\twocolumn[{%
\renewcommand\twocolumn[1][]{#1}%
\maketitle
\centering
\vspace{-5pt}
\includegraphics[width=0.95\linewidth]{sec/figs/teasor.png}
\vspace{-0.87em}
\captionof{figure}{(a) 
While standard convolution and window attention-based methods are constrained by a rigid local window and a fixed connectivity pattern, our GNN-based interactions enable local and global adaptive connections across the image. (b) Rate-Distortion performance and efficiency comparisons on the Tecnick dataset. Upper left is better.
}
\label{fig:teaser}
\vspace{10pt}
}]

\begin{abstract}
Efficient 
\let\thefootnote\relax\footnotetext{
\vspace{-5pt}
* Corresponding author
}
image compression relies on modeling both local and global redundancy. 
Most state-of-the-art (SOTA) learned image compression (LIC) methods are based on CNNs or Transformers, which are inherently rigid. 
Standard CNN kernels and window-based attention mechanisms impose fixed receptive fields and static connectivity patterns, which potentially couple non-redundant pixels simply due to their proximity in Euclidean space. This rigidity limits the model’s ability to adaptively capture spatially varying redundancy across the image, particularly at the global level.
To overcome these limitations, we propose a content-adaptive image compression framework based on Graph Neural Networks (GNNs). Specifically, our approach constructs dual-scale graphs that enable flexible, data-driven receptive fields. Furthermore, we introduce adaptive connectivity by dynamically adjusting the number of neighbors for each node based on local content complexity. These innovations empower our Graph-based Learned Image Compression (GLIC) model to effectively model diverse redundancy patterns across images, leading to more efficient and adaptive compression.
Experiments demonstrate that GLIC achieves state-of-the-art performance, achieving BD-rate reductions of 19.29\%, 21.69\%, and 18.71\% relative to VTM-9.1 on Kodak, Tecnick, and CLIC, respectively.
Code will be released at \url{https://github.com/UnoC-727/GLIC}.

\end{abstract}    

\vspace{-5pt}
\section{Introduction}
\label{sec:intro}
\vspace{-5pt}
With the exponential growth of digital image data, efficient image compression has become indispensable. Lossy compression addresses this need by significantly reducing file sizes while maintaining acceptable visual quality, thereby optimizing storage and transmission efficiency. Recent advancements in learned image compression have demonstrated remarkable potential as an alternative to traditional codecs \cite{balle2016end, balle2018variational, minnen2018joint, minnen2020channel, jiang2023mlic, chen2025knowledgeKDIC, li2025differentiableVMAF, zhang2023neural}. By leveraging advanced nonlinear transforms for encoding and decoding, both CNN-based and transformer-based LIC models have achieved superior rate-distortion performance \cite{liu2023learned, li2025on, jiang2023mlic++}. Due to the quadratic complexity of transformers, recent studies commonly adopt window attention to preserve efficiency \cite{li2023frequency,chen2025s2cformerreorientinglearnedimage}.

However, a fundamental limitation persists: standard CNNs and window attention impose a fixed receptive field and a static connectivity pattern. As illustrated in Fig. \ref{fig:teaser} (a), a $k\times k$ convolution invariably connects each pixel to its fixed $k^2$ neighbors, while window attention also operates within preset blocks. Such rigidity not only prevents the network from expanding its receptive field for distant redundancy, but also tends to couple unrelated features that merely fall inside the same Euclidean neighborhood.
Moreover, natural images exhibit highly non-uniform redundancy: smooth regions carry little information (high redundancy), whereas textured or edge-rich areas contain far more information (low redundancy). Constrained by static connectivity, standard CNN- and Transformer-based LIC models therefore struggle to adapt efficiently to this spatially varying redundancy.

To overcome this limitation, we turn to Graph Neural Networks (GNNs), whose dynamic, data-dependent connectivity naturally captures non-Euclidean relationships beyond fixed Euclidean neighborhoods~\cite{gori2005new,micheli2009neural}. These properties make GNNs a compelling foundation for learned image compression, where redundancy is spatially varying and long-range correlations are common. Building on this insight, we introduce a content-adaptive,  Graph-based Feature Aggregation (GFA) block that fuses two complementary strategies.
First, we design \emph{dual-scale graphs} that jointly leverage a dense local graph and a sparse global graph to capture short- and long-range dependencies with near-linear complexity. By selecting connected nodes from both graphs, the network can flexibly adapt its receptive field across the image, preserving fine details while exploiting distant correlation.
Second, we propose an \emph{adaptive-degree} mechanism that assigns each node a data-driven neighbor quota based on a complexity-aware scoring function, enabling pixel-wise variable neighborhood sizes. This content-adaptive connectivity allocates more capacity to structurally complex regions while avoiding over-connection in smooth areas, improving the model’s ability to identify and eliminate redundancy.

Together, these two strategies yield a simple yet effective route for content-adaptive redundancy removal in LIC models: dual-scale candidate sampling determines the optimal \emph{where} to connect, and adaptive degree determines \emph{how much} to connect. As shown in Fig.~\ref{fig:teaser}(b), integrating our GFA block brings significant performance gains over recent strong baselines while maintaining high efficiency.

Our main contributions are summarized as follows:
\begin{itemize}
\item We construct dual-scale graphs to enable flexible receptive fields, allowing each pixel to select relevant nodes from both neighborhood and distant regions.
\item We propose a complexity-aware scoring mechanism that evaluates compressibility and dynamically determines the optimal connection degree for each node.
\item We present our GNN-based LIC model, GLIC, which outperforms VTM-9.1 by -19.29\%,  -21.69\% and -18.71\% in BD-rate on the Kodak, Tecnick, and CLIC datasets.

\end{itemize}

\section{Related Work}
\subsection{Learned Image Compression}
End-to-end learned image compression has advanced significantly in recent years and some works explore LLMs and Diffusion models for extreme and semantic compression \cite{du2025largeLLM, tian2025smc++, tian2024coding, lu2025vcip, fu2023learned-octaveFU1,  fu2024weconvene, zhou2026dual, ling2026free, liang2025structureGS, chen2026nextframe}. Ball\'{e} \textit{et al.} \cite{balle2016end} pioneered the first CNN-based LIC model, and their subsequent integration of VAEs with hyper-priors \cite{balle2018variational} established a foundational framework. 
Building on this, research has primarily targeted two core components to improve rate-distortion performance: transform networks and entropy models \cite{zafari2023frequency-hilo, mentzer2022vct, ma2019iwave, lu2022transformer, gao2021neural, fu2023asymmetric, begaint2020compressai, zhang2024another, wu2025conditional, llic, akyazi2019learning-anotherwave1}.
Transform design has seen innovations like residual blocks \cite{cheng2020learned}, octave \cite{chen2022two} and invertible networks \cite{xie2021enhanced}. Transformers gained prominence via Swin variants \cite{zou2022devil,zhu2022transformer}, CNN-Transformer hybrids \cite{liu2023learned,chen2025s2cformerreorientinglearnedimage}, and frequency-based attention \cite{li2023frequency}. Some methods \cite{qin2024mambavc, zeng2025mambaic, chen2026contentaware} achieve linear modeling with Mamba.
Entropy modeling advanced with autoregressive \cite{minnen2018joint}, checkerboard \cite{he2021checkerboard}, channel-wise \cite{minnen2020channel} alongside quadtrees \cite{li2022hybrid,li2023neural}, Transformers \cite{qian2022entroformer}, and multi-reference frameworks \cite{jiang2023mlic, jiang2025mlicv2}. T-CA \cite{li2023frequency} further improved channel-wise modeling.

Recent works explore dynamic, content-adaptive architectures for LIC, such as deformable convolutions and graph neural networks (GNNs). Deformable convolutions, applied in image/video compression \cite{dcnic,dcnvc,fu2024fast}, suffer from limited offset ranges. GNN-based methods face challenges: some \cite{tang2022jointgat,gcnic} scale poorly due to quadratic complexity, while others \cite{spadaro2024gabic} are restricted by local receptive fields.

\subsection{Graph Neural Network}
Graph neural networks (GNNs) and graph convolution networks (GCNs) were first introduced in \cite{gori2005new} and \cite{micheli2009neural}, providing a framework for processing graph-structured data. These models have been widely adopted in computer vision tasks, both for 2D and 3D data. In 3D applications, GNNs and GCNs have been applied to point cloud classification, scene graph generation, and action recognition \cite{landrieu2018large, yan2018spatial, xu2017scene, jing2022learning, yang2020distilling}. For 2D tasks, GCNs are effective in image classification, object detection, and semantic segmentation \cite{ma2023image, han2022vision, han2023vision}, where they help capture spatial relationships and contextual dependencies. Additionally, GNNs and GCNs have shown promise in low-level vision tasks such as image denoising and super-resolution, where images are treated as graphs to enhance performance \cite{liu2022dual, zhou2020cross, tian2024image, li2021cross, mou2021graph}.

\begin{figure*}
\centering
\includegraphics[width=1\textwidth]{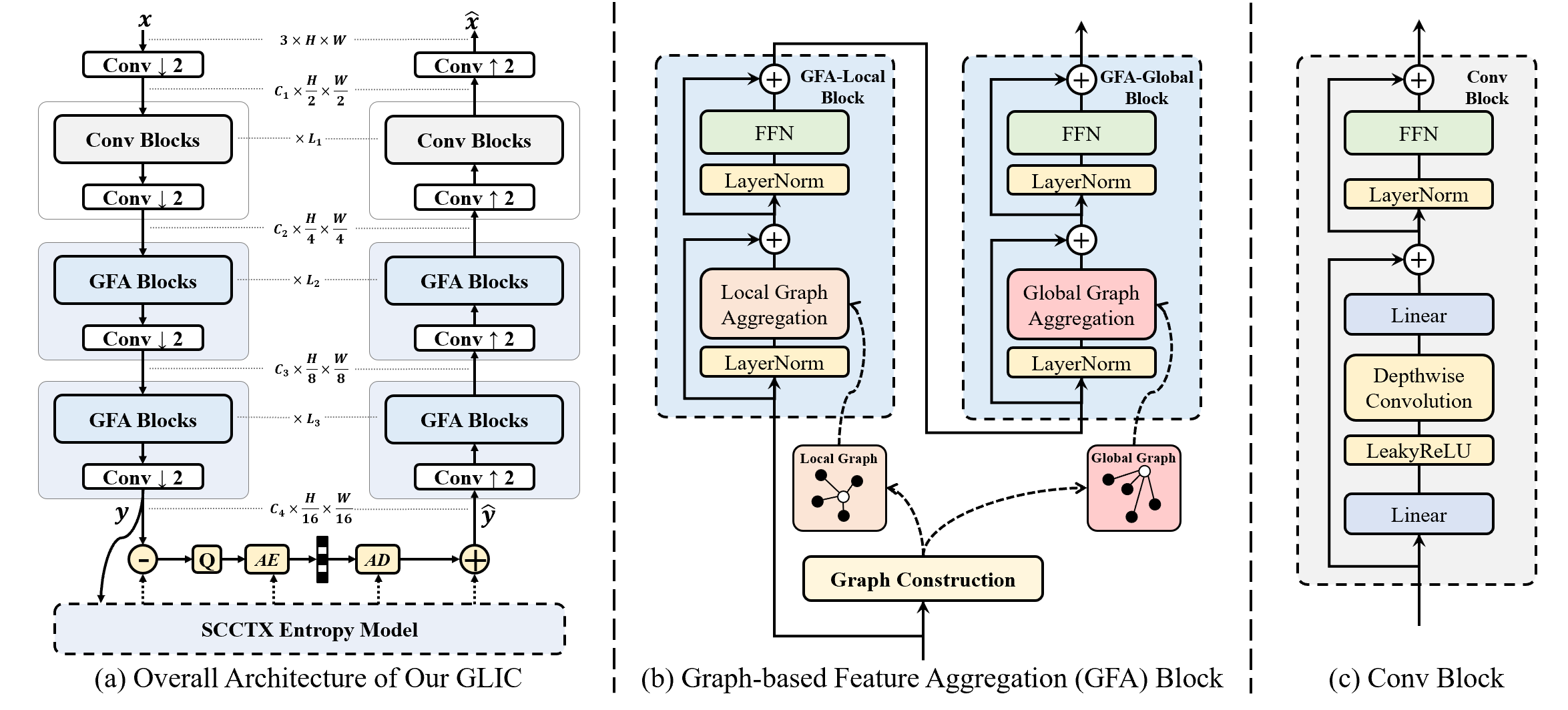}
\vspace{-15pt}
\caption{\textbf{Overview of our method.} (a) Architecture of the proposed GLIC codec. Channel widths are {$C_1,C_2,C_3,C_4$}, and the numbers of non-linear transform blocks are {$L_1,L_2,L_3$}.
(b) Graph-based Feature Aggregation Block used as advanced non-linear transforms.
(c) Lightweight Conv Block for early feature extraction and reduced complexity.
} 
\vspace{-10pt}
\label{backbone} 
\end{figure*}

\vspace{-3pt}
\section{Method}
\vspace{-3pt}

\subsection{Preliminary}
\vspace{-3pt}

Fig. \ref{backbone} (a) shows a standard LIC pipeline with an \textit{Encoder} $g_a$, a \textit{Decoder} $g_s$, and an \textit{Entropy Model}.  
The encoder maps an RGB image $\bm{x}$ to a latent $\bm{y}$, which is quantized into $\hat{\bm{y}}$ and decoded to the reconstruction $\hat{\bm{x}}$:  
\[
\bm{y}=g_a(\bm{x};\bm{\theta}_a),\;
\hat{\bm{y}}=Q(\bm{y}-\bm{\mu})+\bm{\mu},\;
\hat{\bm{x}}=g_s(\hat{\bm{y}};\bm{\theta}_s).
\]

To estimate the entropy parameters $(\bm{\mu},\bm{\sigma})$, a \textit{Hyper Encoder-Decoder} pair converts $\bm{y}$ to side-information $\hat{\bm{z}}$:  
\[
\bm{z}=h_a(\bm{y};\bm{\phi}_a),\;
\hat{\bm{z}}=Q(\bm{z}),\;
(\bm{\mu},\bm{\sigma})=h_s(\hat{\bm{z}};\bm{\phi}_s).
\]

Training minimizes the rate-distortion objective  
\[
\mathcal{L}= \mathcal{R}(\hat{\bm{y}})+\mathcal{R}(\hat{\bm{z}})+\lambda\,\mathcal{D}(\bm{x},\hat{\bm{x}}),
\]
where $\mathcal{R}(\cdot)$ is the predicted bitrate, $\mathcal{D}(\cdot,\cdot)$ the distortion metric, and $\lambda$ balances rate against reconstruction quality.

\vspace{-3pt}
\subsection{Overview of Our Method}
\vspace{-3pt}
As shown in Fig.~\ref{backbone}, our Graph-based Learned Image Compression (GLIC) model adopts a standard VAE-based LIC backbone. We divide the nonlinear transforms into three stages by spatial-resolution and deploy our \emph{Graph-based Feature Aggregation} (GFA) block in the latter two stages. Within each stage, features are aggregated sequentially by GFA-Local and  GFA-Global blocks with dual-scale graphs.

At a high level, each GFA stage (i) forms a dual-scale candidate set per node to realize flexible receptive fields; (ii) assigns a  complexity-aware degree budget driven by an RMS-Gradient score; (iii) constructs a directed graph by per-node thresholding with a bisection search; and (iv) performs edge-conditioned aggregation for feature updating.

\vspace{-3pt}
\subsection{Graph-based Feature Aggregation (GFA) Block}
\vspace{-3pt}
\label{sec:gfa}
Let $F\in\mathbb{R}^{H\times W\times C}$ be a feature map with $N=H\!\times\! W$ pixels. We index feature pixels (and thus graph nodes) by $i\in\{1,\dots,N\}$ and denote the one-to-one coordinate mapping by $\pi(i)=(u_i,v_i)$. Each node is associated with a $C$-dimensional feature vector $x_i\in\mathbb{R}^C$.
The proposed GFA block first fixes a global connectivity budget, specified by a target average in-degree $\bar d$ or equivalently a total edge budget $B = N\bar d$. This budget is then allocated to individual nodes according to their local complexity, yielding node-wise quotas $q_i$ and target degrees $d_i^\star$. Based on these targets, each node selects its neighbors from a dual-scale candidate set that combines local and global samples, after which feature aggregation is performed on the resulting directed graph. Therefore, the proposed design separates \emph{where to connect} (dual-scale candidate sampling) from \emph{how much to connect} (complexity-aware degree budgeting).

The GFA workflow comprises four steps:
\begin{enumerate}
\item \textbf{Dual-scale candidate sampling} (\S\ref{sec:dual-scale}): for each node $i$, we construct a  local candidate set $\mathcal{C}_i^{\mathrm{loc}}$ and a  global candidate set $\mathcal{C}_i^{\mathrm{glo}}$, yielding a flexible receptive field.
\item \textbf{Complexity-aware degree budgeting} (\S\ref{sec:complexity}): each node is assigned a \emph{neighbor quota} $q_i$ that specifies its target in-degree.
\item \textbf{Graph construction} (\S\ref{sec:graph-construction}): given $\mathcal{C}_i$ and $q_i$, node $i$ selects its connected neighbors via per-node thresholding on cosine similarity, resulting in a directed adjacency $A$.
\item \textbf{Graph aggregation} (\S\ref{sec:graph-aggregation}): features are updated by edge-conditioned aggregation with dual-graphs.
\end{enumerate}

We model the feature map as a \emph{directed} graph $\mathcal{G}=(\mathcal{V},\mathcal{E})$, where $\mathcal{V}=\{1,\dots,N\}$ refers to vertices and a directed edge $j\!\to\! i$ indicates that $j$ contributes a message to the update of $i$. For efficiency, each node first constructs a candidate set $\mathcal{C}_i\subseteq\mathcal{V}$ and then selects a subset $\mathcal{N}_i\subseteq\mathcal{C}_i$ with size $d_i\triangleq|\mathcal{N}_i|$ for feature aggregation.

\subsubsection{Flexible Receptive Fields via Dual-scale Sampling}
\label{sec:dual-scale}
The spatial footprint of candidate set $\mathcal{C}_i$ shapes the receptive field of node $i$. To avoid $O(N^2)$ global pairwise computations of similarities to select  neighbors, we adopt a \emph{dual-scale} sampling scheme whose per-node cost is near-linear in the number of samples, while approximately expanding the effective receptive field to the full image.

\noindent\textbf{Dense local sampling.}
We collect candidates within a fixed-size neighborhood centered, similar to CNN and window attention,  which captures fine-grained local structure crucial for low-level vision.
Let $L\in\mathbb{N}$ be the local window size. For $\pi(i) = (u_i, v_i)$, we define the local candidate set as
\begin{equation}
\begin{split}
\mathcal{C}^{\mathrm{loc}}_i 
&= \bigl\{\, j \in \mathcal{V}\; \big|\; \pi(j) = (u_i + \Delta_u,\, v_i + \Delta_v), \\
&\quad \Delta_u,\Delta_v \in \mathbb{Z},\; |\Delta_u|,|\Delta_v|\le \tfrac{L-1}{2}\,\bigr\},
\end{split}
\end{equation}

\noindent\textbf{Sparse global sampling.}
We introduce a mesh-grid sampling scheme inspired by dilated convolutions and stride sampling \cite{DBLP:journals/corr/YuK15,hassani2023dilatedneighborhoodattentiontransformer,li2021cross,zhou2020cross}. Unlike the approach in \cite{tian2024image} that relies on a fixed set of global anchors, we generate per-node mesh-grids, activating more pixels for global interactions.

Let $G\in\mathbb{N}$ be the grid size. Define strides
$\Omega_h=\big\lfloor H/G \big\rfloor$ and $\Omega_w=\big\lfloor W/G \big\rfloor$.
We place a $G\times G$ mesh centered at $\pi(i)$ by recording its position within the stride lattice via
$r_i \equiv u_i \bmod \Omega_h$ and $c_i \equiv v_i \bmod \Omega_w$.
The global candidate set is the lattice points aligned with $(r_i,c_i)$:
\begin{equation}
\begin{split}
\mathcal{C}^{\mathrm{glo}}_i 
&= \bigl\{\, j\in\mathcal{V}\; \big|\; \pi(j)=(r_i+u\,\Omega_h,\, c_i+v\,\Omega_w), \\
&\quad u\in\{0,\dots,G-1\},\; v\in\{0,\dots,G-1\}\,\bigr\},
\end{split}
\end{equation}

Dense local sampling captures fine structures while sparse global sampling contributes long-range context, yielding an adaptive receptive field without quadratic cost.

\vspace{-3pt}
\subsubsection{Complexity-aware Neighbor Quota Assignment}
\vspace{-3pt}
\label{sec:complexity}

Node degrees need not be fixed in a GNN. We allocate each node a target in-degree (quota) proportional to a pixel-wise complexity score. Intuitively, pixels with larger local variation are harder to compress and should attend to more neighbors to aggregate information for removing redundancy. We propose to use Sobel operator \cite{sobel19683x3} to calculate gradients and assess the local complexity of each feature pixel.

\noindent\textbf{RMS-Gradient (RMS-G) complexity scoring.}
Using the $3{\times}3$ Sobel filters $K_x,K_y$, we compute per-channel gradients
\begin{equation}
S_{x,c}=F_c * K_x,\quad S_{y,c}=F_c * K_y,\qquad c=1,\dots,C,
\end{equation}
and define the magnitude and the cross-channel Root Mean Square (RMS) score at coordinate $(u,v)$ as
\begin{align}
\bigl|\nabla F_c(u,v)\bigr| &= \sqrt{S_{x,c}(u,v)^2 + S_{y,c}(u,v)^2},\\
s(u,v) &= \sqrt{\tfrac{1}{C}\sum_{c=1}^{C} \bigl|\nabla F_c(u,v)\bigr|^2 }.
\end{align}
Let $s_i \triangleq s\!\bigl(\pi(i)\bigr)\ge 0$. Compared with down/up-sampling residual proxies \cite{tian2024image}, Sobel directly measures first-order spatial variation and emphasizes large local changes (edges, fine details), which typically correlate with coding difficulty. Moreover, since RMS is sensitive to larger values, it better captures the energy compaction phenomenon \cite{li2025on} compared to the simple averaging strategy used by \cite{tian2024image}.  Further discussion of this choice is provided in the supplementary material.

\noindent\textbf{From global budget to per-node quotas.}
Let $\bar d$ be the target \emph{average} in-degree and $B \triangleq N\,\bar d$ the total edge budget. We convert $\{s_i\}$ into normalized weights
$
w_i \;=\; \frac{s_i}{\sum_{k=1}^{N} s_k},
$
and allocate real-valued quotas $q_i = B\,w_i$. The target degree is
\begin{equation}
d_i^\star \;=\; \mathrm{clip}\bigl(\mathrm{round}(q_i),\;1,\; n_i \bigr),
\end{equation}
ensuring at least one neighbor and at most all candidates.

\vspace{-3pt}
\subsubsection{Graph Construction}
\vspace{-3pt}
\label{sec:graph-construction}
The primary redundancy lies among content-similar pixels. To promote redundancy removal, each node should connect to content-correlated candidates by thresholding cosine similarity. For node features $x_i,x_j\in\mathbb{R}^C$, define
\begin{equation}
S_{ij} \;=\; \frac{\langle x_i, x_j\rangle}{\|x_i\|_2\,\|x_j\|_2}, \qquad j\in\mathcal{C}_i.
\end{equation}
We seek a per-node threshold $\theta_i$ such that the selected count
\begin{equation}
m_i(\theta_i) \;\triangleq\; \#\bigl\{\, j\in\mathcal{C}_i \,:\, S_{ij}\ge \theta_i \,\bigr\}
\;\approx\; d_i^\star,
\end{equation}
where $\#\{\cdot\}$ denotes set cardinality. The resulting directed neighborhood and determined adjacency are
\begin{equation}
\mathcal{N}_i \;=\; \bigl\{\, j \in \mathcal{C}_i \;:\; S_{ij} \ge \theta_i \,\bigr\}, A_{ij} \;=\;
\begin{cases}
1, & \text{if } j \in \mathcal{N}_i,\\
0, & \text{otherwise.}
\end{cases}
\end{equation}

\noindent\textbf{Per-node bisection for $\theta_i$.}
We solve $m_i(\theta_i)\!\approx\! d_i^\star$ via a few iterations of bisection. Let
$\ell_i=\min_{j\in\mathcal{C}_i} S_{ij}$, $u_i=\max_{j\in\mathcal{C}_i} S_{ij}$, we initialize $\theta_i=\tfrac{1}{n_i}\sum_{j\in\mathcal{C}_i} S_{ij}$. Each iteration halves the feasible interval, yielding near-linear cost in $n_i$. Detailed algorithm is provided below.

\begin{algorithm}[H]
\caption{Per-node threshold searching via bisection}
\label{alg:bisection}
\begin{algorithmic}[1]
\State \textbf{Input:} features $\{x_i\}_{i=1}^N$, candidates $\{\mathcal{C}_i\}$, targets $\{d_i^\star\}$, iterations $T$
\State \textbf{Output:} adjacency $A\in\{0,1\}^{N\times N}$
\State Compute $S_{ij}$ for all $i$ and $j\!\in\!\mathcal{C}_i$ \hfill (cosine similarity)
\For{each node $i$ \textbf{in parallel}}
    \State $\ell_i \leftarrow \min_{j\in\mathcal{C}_i} S_{ij}$;\quad
           $u_i \leftarrow \max_{j\in\mathcal{C}_i} S_{ij}$;\quad
           $\theta_i \leftarrow \tfrac{1}{|\mathcal{C}_i|}\sum_{j\in\mathcal{C}_i} S_{ij}$
    \For{$t=1$ \textbf{to} $T$}
        \State $m_i \leftarrow \#\{\, j\in\mathcal{C}_i : S_{ij}\ge \theta_i \,\}$
        \If{$m_i > d_i^\star$}
            \State $\ell_i \leftarrow \theta_i$
        \Else
            \State $u_i \leftarrow \theta_i$
        \EndIf
        \State $\theta_i \leftarrow (\ell_i + u_i)/2$
    \EndFor
    \State $\mathcal{N}_i \leftarrow \{\, j\in\mathcal{C}_i : S_{ij}\ge \theta_i \,\}$;\quad
           set $A_{ij}\!\leftarrow\!1$ for $j\in\mathcal{N}_i$ and $A_{ij}\!\leftarrow\!0$ otherwise
\EndFor
\end{algorithmic}
\label{algot1}
\end{algorithm}

\begin{table*}[t]
\centering
  \scriptsize 
  \setlength{\tabcolsep}{3.5pt} 
  \renewcommand{\arraystretch}{1.} 

\caption{\textbf{Connectivity-oriented comparison.}
Notation: $N$ denotes the total number of nodes/pixels; $k$ the side length of the convolution kernel; $L$ the side length of the attention/GNN window; and $K$ the average neighborhood size (connectivity and in-degrees for GNNs). Computational cost is reported as \emph{construction + aggregation}. ``Construction'' counts candidate search and similarity calculation.}
\vspace{-6pt}

\label{tab:unified_connectivity}
\renewcommand{\arraystretch}{1.1}
\setlength{\tabcolsep}{5pt}
\begin{tabularx}{\textwidth}{l X X X l}
\toprule
\textbf{Method}
& \textbf{Candidates}
& \textbf{Selection \& Degree}
& \textbf{Receptive Field}
& \textbf{Complexity} \\
\midrule
\multicolumn{5}{l}{\emph{Canonical operators}} \\ 

Dynamic CNN
& $k{\times}k$ window
& keep-all; degree $=k^2$ 
& bounded by window
& $\mathcal{O}(1)\;+\;\mathcal{O}(N k^2)$ \\

Deformable CNN
& offset-enlarged $k{\times}k$ window
& keep-all; degree $=k^2$
& offset-enlarged local window
& $\mathcal{O}(N k)\;+\;\mathcal{O}(N k)$ \\

Window Attention
& $L{\times}L$ window
& keep-all; degree $=L^2$
& bounded by window 
& $\mathcal{O}(1)\;+\;\mathcal{O}(N L^2)$ \\

Transformer (global)
& all nodes
& keep-all; degree $=N$
& full
& $\mathcal{O}(1)\;+\;\mathcal{O}(N^2)$ \\

\midrule
\multicolumn{5}{l}{\emph{GNN-based LIC}} \\
\midrule
Fully-connected GAT \cite{tang2022jointgat}
& all nodes
& keep-all; degree $=N{-}1$ 
& full
& $\mathcal{O}(1)\;+\;\mathcal{O}(N^2)$ \\

Global $k$NN GNN \cite{gcnic}
& all nodes
& $k$NN; degree $=K$ 
& full
& $\mathcal{O}(N^2)\;+\;\mathcal{O}(N K)$ \\

Window $k$NN GAT \cite{spadaro2024gabic}
& $L{\times}L$ window
& $k$NN; degree $=K$ 
& bounded by window
& $\mathcal{O}(N L^2)\;+\;\mathcal{O}(N K)$ \\

\textbf{GLIC (ours)}
& local window $\cup$ sparse global grid (n candidates in total)
& per-node threshold and adaptive degree quota; average degree $K$
& near-full 
& $\mathcal{O}(N n)\;+\;\mathcal{O}(N K)$ \\
\bottomrule
\end{tabularx}
\vspace{-10pt}
\label{tab:unified_connectivity}
\end{table*}

\noindent\textbf{Complexity.}\;
Per node, similarity evaluation and bisection over $T$ iterations cost
$O(n_i + T\log n_i)$ time; in practice with small $T$ and moderate $n_i\!\approx\! L^2{+}G^2$, this is near-linear.

\vspace{-3pt}
\subsubsection{Graph Aggregation}
\label{sec:graph-aggregation}
\vspace{-3pt}
Following edge-conditioned aggregation~\cite{simonovsky2017dynamic,zhou2020cross,tian2024image}, we update node features over the constructed directed graphs. Let $z_i^{\,\ell}\in\mathbb{R}^{C_\ell}$ be the feature of node $i$ at layer $\ell$. We compute
\begin{align}
z_i^{\,\ell}
&= \sum_{j\in \mathcal{N}_i}
\alpha_{i\leftarrow j}^{\,\ell}\; \phi^{\,\ell}\!\bigl(z_j^{\,\ell-1}\bigr),
\label{eq:z-update} \\[2pt]
\alpha_{i\leftarrow j}^{\,\ell}
&= \frac{\exp\bigl(g^{\,\ell}(z_i^{\,\ell-1}, z_j^{\,\ell-1})\bigr)}
{\sum_{u\in\mathcal{N}_i}\exp\bigl(g^{\,\ell}(z_i^{\,\ell-1}, z_u^{\,\ell-1})\bigr)}.
\label{eq:alpha-def}
\end{align}
where $\phi^{\,\ell}$ is a learnable linear projection  and
$g^{\,\ell}$ is cosine similarity.
This design prioritizes pairwise pixel relations and preserves contextual information across both analysis and synthesis transforms.

\vspace{-3pt}
\subsection{Connectivity Comparisons}
\vspace{-3pt}
In this section we compare the connectivity pattern and complexity of canonical operators and prior
GNN-based LIC models (Table~\ref{tab:unified_connectivity}). Dynamic and deformable CNNs as well as
window-based attention and window-based GAT restrict connections to local or window-bounded regions,
thereby missing truly global redundancy, while global Transformers and fully-connected GATs achieve
full receptive fields at the cost of quadratic complexity $\mathcal{O}(N^2)$. In contrast, our GLIC
constructs dual-scale graphs with a small candidate set and adaptive degree, attaining a near-global
receptive field while keeping both graph construction and aggregation linear in $N$ (i.e., $\mathcal{O}(Nn)$
and $\mathcal{O}(NK)$), which achieves near-full global
connectivity without quadratic overhead.

\vspace{-3pt}
\section{Experiments}
\vspace{-3pt}
\label{sec:Experiments}

\subsection{Experimental Setup}
\vspace{-3pt}
\subsubsection{Training Details}
\vspace{-3pt}
We train GLIC models on the Flickr2W dataset \cite{liu2020unified}. Models are optimized with Adam \cite{kingma2014adam} with an initial learning rate of 0.0001. We adopt the training acceleration strategy introduced by \cite{li2025on}, reducing the total training time to 3 days.
To measure the distortion, we employ mean square error (MSE) and multiscale structural similarity (MS-SSIM).
For MSE-optimized models, we set Lagrangian multipliers $\lambda$ as \{0.0017, 0.0035, 0.0067, 0.0130, 0.0250, 0.050\}; for MS-SSIM-optimized models, we set \{3, 5, 8, 16, 36, 64\}. 
All experiments are conducted on NVIDIA A100 GPUs.

\begin{figure*}[t]
    \centering
    \begin{minipage}{0.49\linewidth}
        \centering
        \includegraphics[width=1\linewidth]{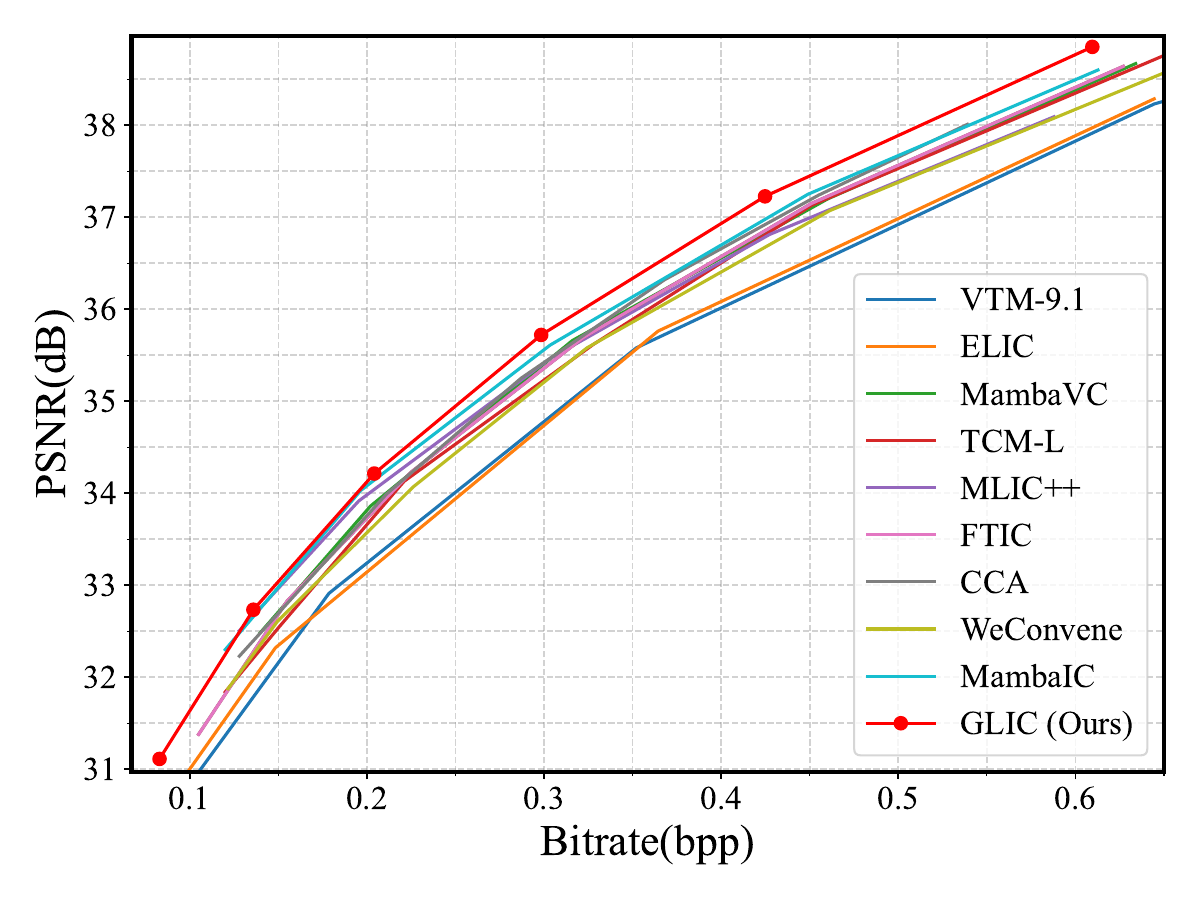}
        \captionsetup{skip=5.5pt} 
        \vspace{-25pt}
        \caption{PSNR R-D curves on the CLIC 2020 dataset.}
        \label{fig:kodak_psnr}
    \end{minipage}
    \hfill
    \begin{minipage}{0.49\linewidth}
        \centering
        \includegraphics[width=1\linewidth]{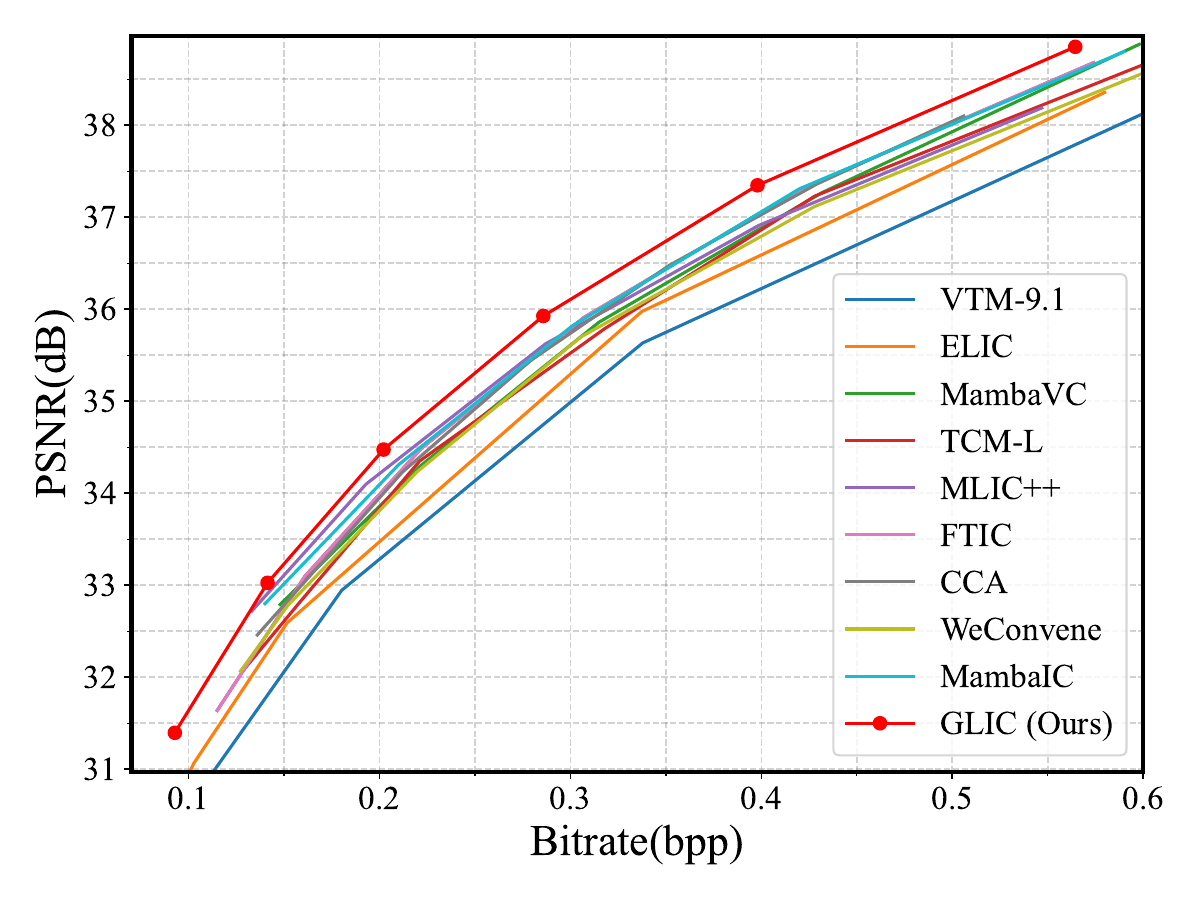}
        \captionsetup{skip=5.5pt} 
        \vspace{-25pt}
        \caption{PSNR R-D curves on the Tecnick dataset.}
        \label{fig:tecnick_psnr}
    \end{minipage} 
    \label{rdcurves2}
\end{figure*}
\begin{figure*}[t]
\vspace{-13pt}
    \centering
    \begin{minipage}{0.49\linewidth}
        \centering
        \includegraphics[width=1\linewidth]{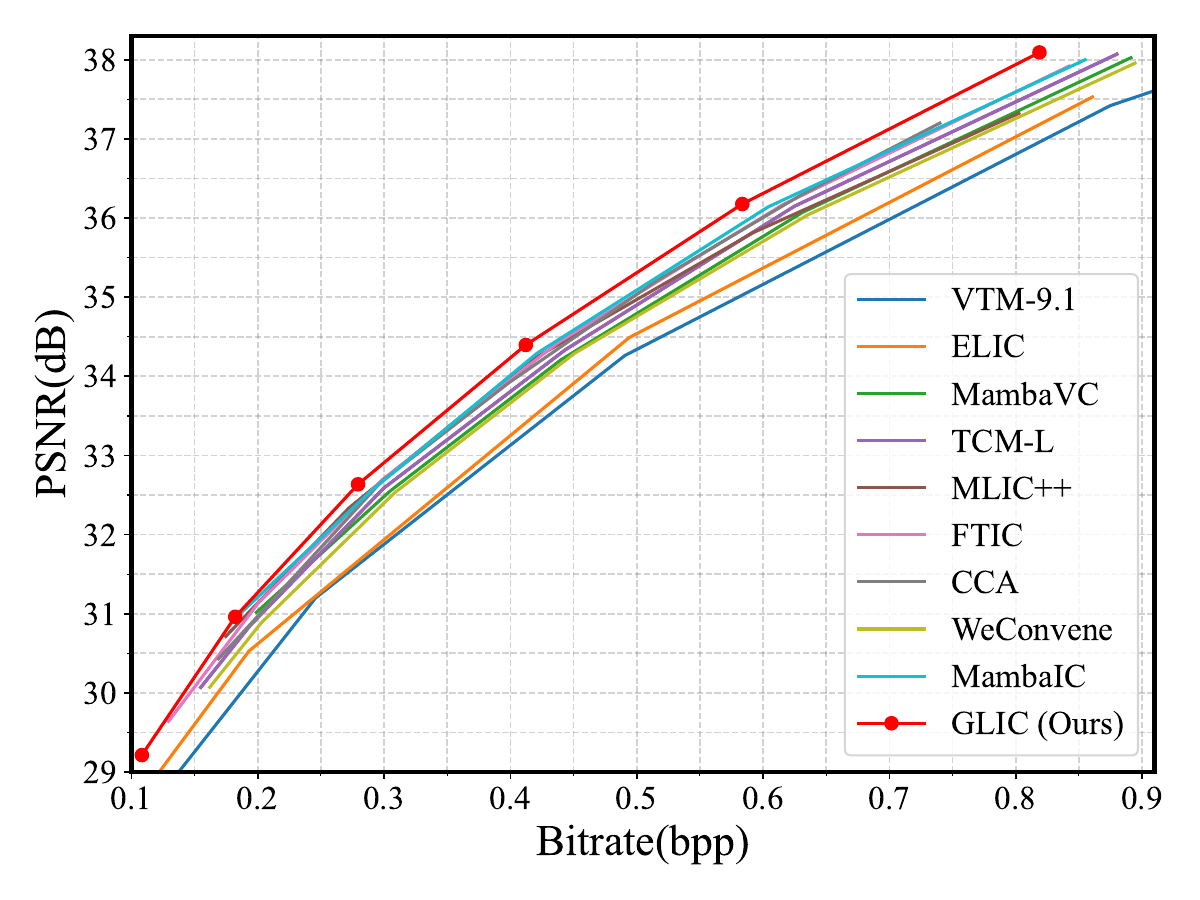}
        \captionsetup{skip=2.5pt} 
        \label{fig:clic_psnr}
    \end{minipage}
    \hfill
    \begin{minipage}{0.49\linewidth}
        \centering
        \includegraphics[width=1\linewidth]{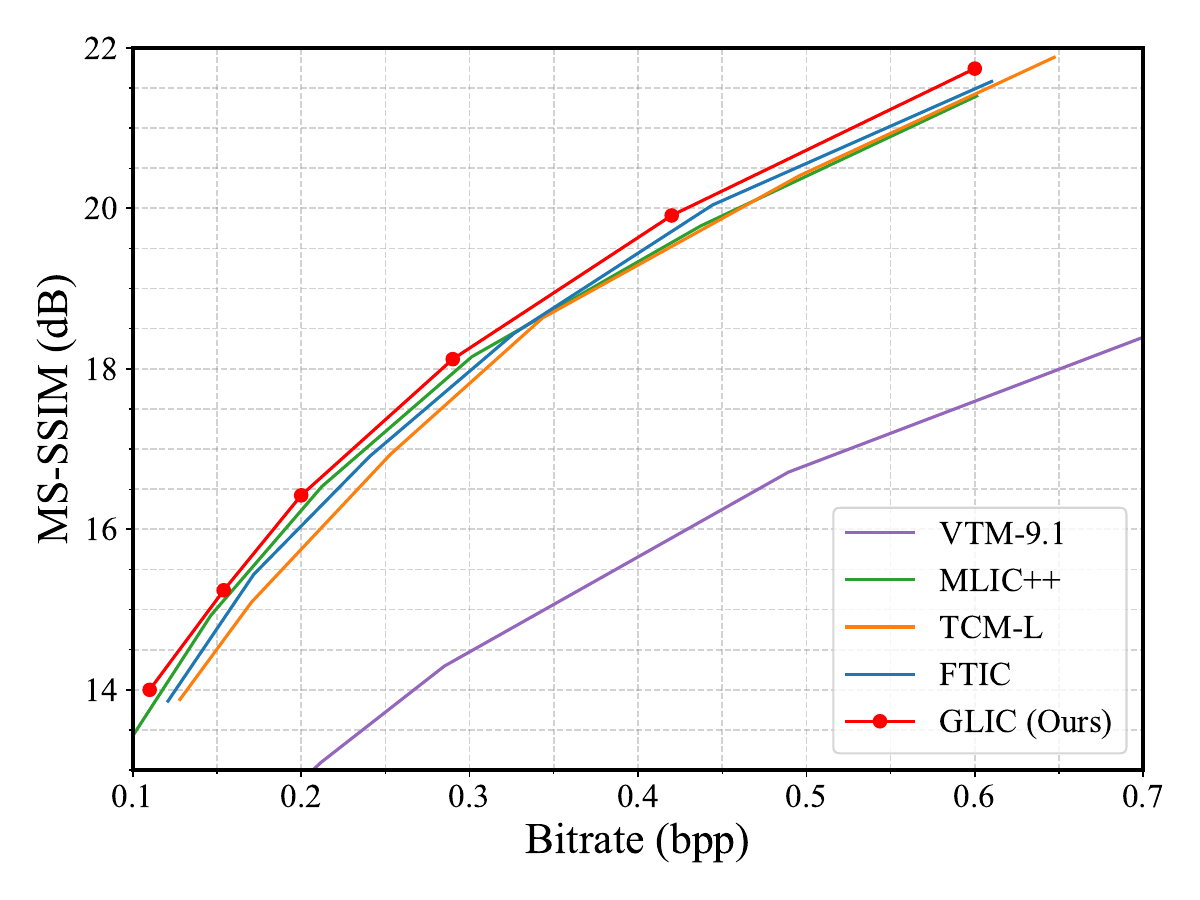}
        \captionsetup{skip=2.5pt} 
        \label{fig:clic_msssim}
    \end{minipage} 
        \captionsetup{skip=2.5pt} 
        \vspace{-20pt}
    \caption{PSNR and MS-SSIM R-D curves on the Kodak Dataset.}
    \label{clic} 
    \vspace{-15pt}
\end{figure*}

\vspace{-3pt}
\subsubsection{Evaluation}
\vspace{-3pt}

\begin{table*}[t]
  \centering
  \setlength{\tabcolsep}{4pt}
  \renewcommand{\arraystretch}{1.1}
  \fontsize{8.7}{10}\selectfont
  \caption{\textbf{Quantitative Comparisons of leading LIC methods.} BD‑rates are evaluated against VTM‑9.1.  FLOPs, peak memory and decoding time are measured on 2K‑resolution images.}
\vspace{-8pt}
  \begin{tabular}{l|ccccc|ccc}
    \hline
    \multirow{2}{*}{\textbf{Method}} &
      \multicolumn{5}{c|}{\textbf{Complexity}} &
      \multicolumn{3}{c}{\textbf{BD‑rate (\%) $\downarrow$}} \\ 
    \cline{2-9}
      & \textbf{Params (M)} & \textbf{FLOPs (T)} & \textbf{Enc‑Lat. (s)} &
        \textbf{Dec‑Lat. (s)} & \textbf{Peak‑Mem. (G)} &
        \textbf{Kodak} & \textbf{Tecnick} & \textbf{CLIC} \\ 
    \midrule
    VTM‑9.1 \cite{bross2021overviewvvc}   &   -   &   -   &   -  & - &  -  &  0.00  &  0.00  &  0.00  \\
    ELIC (CVPR'22)  \cite{he2022elic}    & 33.29 & 1.74 & 0.583 & 0.335 & 1.50 & -5.95  & -7.68  & -1.20  \\
    MLIC++ (ICML'23W) \cite{jiang2023mlic++}    & 116.48  & 2.64 & 0.508 & 0.547 & 2.08 & -15.14 & -17.23 & -14.41 \\
    TCM-L (CVPR'23)   \cite{liu2023learned}    & 75.89 & 3.74 & 0.647 & 0.542 & 7.73 & -13.42 & -10.93 &  -9.10 \\
    FTIC  (ICLR'24) \cite{li2023frequency}    & 69.78 & 2.38 & $>$10 & $>$10 & 4.90 & -14.83 & -14.39 & -10.70 \\
    CCA (NeurIPS'24)   \cite{han2024causal}    & 64.89 & 3.28 & 0.526 & 0.385 & 5.04 & -13.94 & -14.13 & -11.93 \\
    WeConvene (ECCV'24) \cite{fu2024weconvene} & 105.51 & 4.82 & 1.264 & 1.293 & 4.53 & -8.96  & -10.70 & -7.55  \\
    HPCM  (ICCV'25)  \cite{iccv25HPCM}  & 68.50 & 2.00 & 0.532 & 0.498 & 5.89 & -16.13 & -17.26  & -15.02 \\
    DCAE (CVPR'25)   \cite{cvpr2025diction_dcae}  & 119.22 & 2.28 & 0.428 & 0.449 & 5.59 & -17.18 & -20.07 & -16.91 \\
    LALIC (CVPR'25)   \cite{lalic}  & 63.24 & 2.53 & 0.779 & 0.362 & 3.89 & -15.50 & -17.71 & -15.47 \\
    MambaIC (CVPR'25) \cite{zeng2025mambaic}  & 157.09   &  5.56 & 1.436  &   0.669   &  20.32   & -15.13 & -15.78 & -15.73 \\
    \hline
    \textbf{GLIC (Ours)} & 67.20 & 2.48 & 0.617 & 0.395 & 5.46 & \textbf{-19.29} & \textbf{-21.69} & \textbf{-18.71} \\
    \hline
  \end{tabular} 
  \vspace{-8pt}
  \label{tab:bdrates_complexity} 
\end{table*}

We evaluate our models on three datasets: the Kodak image set \cite{kodak1993} with a resolution of 768 × 512, the Tecnick test set \cite{asuni2014testimages} with a resolution of 1200 × 1200, and the CLIC2020Val dataset \cite{CLIC2020} with 2K resolution. For RD performance evaluation, we employ PSNR, MS-SSIM, and bits per pixel (bpp) as the key metrics. We use BD-rate \cite{bjontegaard2001} to measure the average bitrate savings. We also use BD-PSNR to measure the average PSNR gain at the same bitrates. For model complexity, we utilize parameter count, FLOPs, encoding/decoding latency and peak memory for 2K images as metrics.

\vspace{-3pt}
\subsubsection{Model Details}
\vspace{-3pt}
We set channel numbers \{$C_1, C_2, C_3, C_4$\} as \{128, 192, 192, 320\} and transform block numbers \{$L_1, L_2, L_3$\} as \{3, 5, 5\}. We employ SCCTX model \cite{he2022elic} with 5 slices for entropy coding. The dense sampling size is set as 8 and sparse sampling size is 16. We set the average node degree as 64.

\vspace{-3pt}
\subsection{Rate-Distortion Performance}
\vspace{-3pt} 
We benchmark our method against the traditional image codec VTM-9.1 \cite{bross2021overviewvvc}, as well as several SOTA LIC methods including  FTIC \cite{li2023frequency}, CCA \cite{han2024causal}, WeConvene \cite{fu2024weconvene}, HPCM \cite{iccv25HPCM}, LALIC \cite{lalic}, DCAE \cite{cvpr2025diction_dcae}, MambaIC \cite{zeng2025mambaic}. The resulting RD curves, depicted in Fig.~\ref{fig:kodak_psnr}-\ref{clic}, illustrate the performance trends across different bit rates. Detailed BD-rate comparisons, with VTM-9.1 as the anchor, are reported in Tab.~\ref{tab:bdrates_complexity}. Our GLIC model outperforms VTM-9.1 by -19.29\%, and -21.69\%, -18.71\% in BD-rate on Kodak, Tecnick, and CLIC datasets, significantly surpassing previous methods.

Specifically, in terms of BD-PSNR improvements, when compared with FTIC \cite{li2023frequency}, our method achieves gains of  0.26 dB, 0.37 dB, and 0.38 dB on the Kodak, CLIC, and Tecnick datasets, respectively. Furthermore, the improvements become even more pronounced against TCM-L \cite{liu2023learned}, with BD-PSNR increases of 0.39 dB, 0.46 dB, and 0.56 dB on the corresponding datasets. In terms of MS-SSIM, our GLIC model also achieves  0.46 dB and 0.28 dB improvements on the CLIC dataset compared to TCM-L and FTIC, respectively, as shown in Fig. \ref{clic}.
These results not only demonstrate the robustness of our approach but also validate its superiority across a variety of resolutions.

\begin{figure*}[t]
\centering
\includegraphics[width=\textwidth]{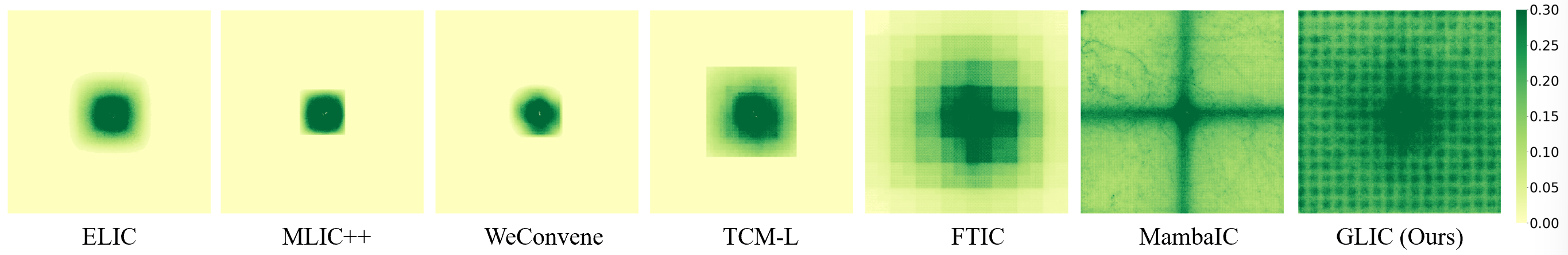}\\[-2pt]
\includegraphics[width=\textwidth]{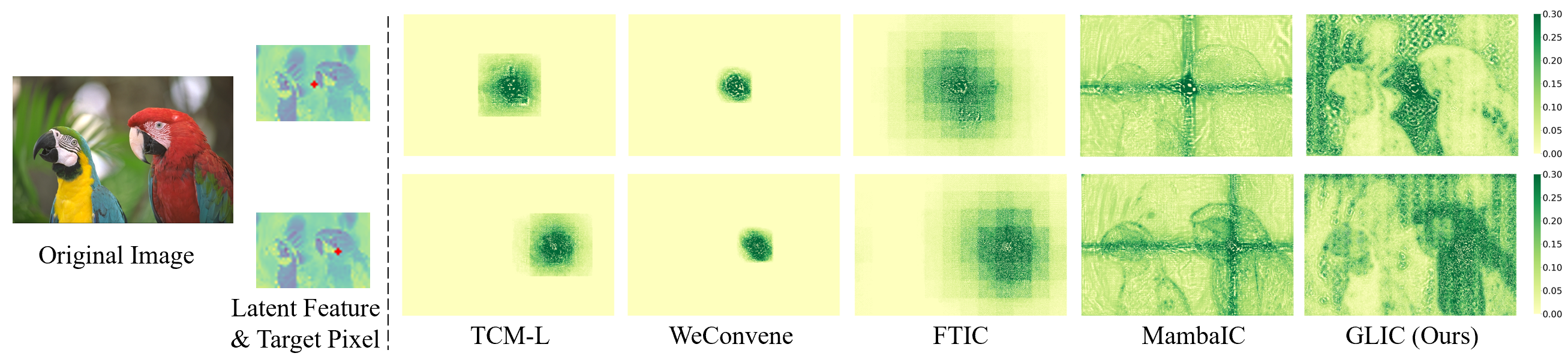}
\captionsetup{skip=1pt}
\caption{(Top) Averaged Effective Receptive Fields of state-of-the-art LIC models across the 24 Kodak images. (Bottom) Single-image Effective Receptive Field for Kodak image 23, demonstrating the adaptivity of our GLIC model. More results are in the supplementary.}
\label{fig:erf_combo}
\vspace{-12pt}
\end{figure*}

\vspace{-3pt}
\subsection{Effective Receptive Fields}
\vspace{-3pt}
We visualize the effective receptive fields (ERFs) \cite{luo2016understanding} of LIC models to demonstrate the superiority of our method. We first show the averaged ERF on the Kodak dataset in Fig.~\ref{fig:erf_combo}. In these visualizations, a larger ERF appears as a broader spread of dark regions. Compared with prior methods, our GLIC model exhibits a substantially larger ERF, attributable to its dual-graph strategy, which adaptively captures long-range dependencies within images.

\noindent\textbf{Single-image ERF.} We also present single-image ERF visualizations (Fig.~\ref{fig:erf_combo}) using Kodak image 23. ERFs are computed from two locations in the latent feature map: the center and a point slightly to its right (details of the ERF computation are provided in the supplementary material). Although these points are close in space, one lies in the background and the other on the foreground parrot. Prior models produce ERFs with similar, mildly isotropic shapes at both locations. In contrast, our method yields clearly different, content-aware ERFs: for the central background point, the ERF concentrates on background regions (especially the blades of grass), largely skipping the nearby parrot. For the foreground point on the parrot, despite its Euclidean proximity to the first point, the ERF focuses on the red feathers of the right parrot. These results align with human perception of redundancy and semantic structure, indicating that our model performs content-adaptive redundancy detection and elimination. This is a dynamic behavior absent from existing methods, highlighting the advantage of our approach.

\begin{figure*}[ht]
\centering
\includegraphics[width=0.9\textwidth]{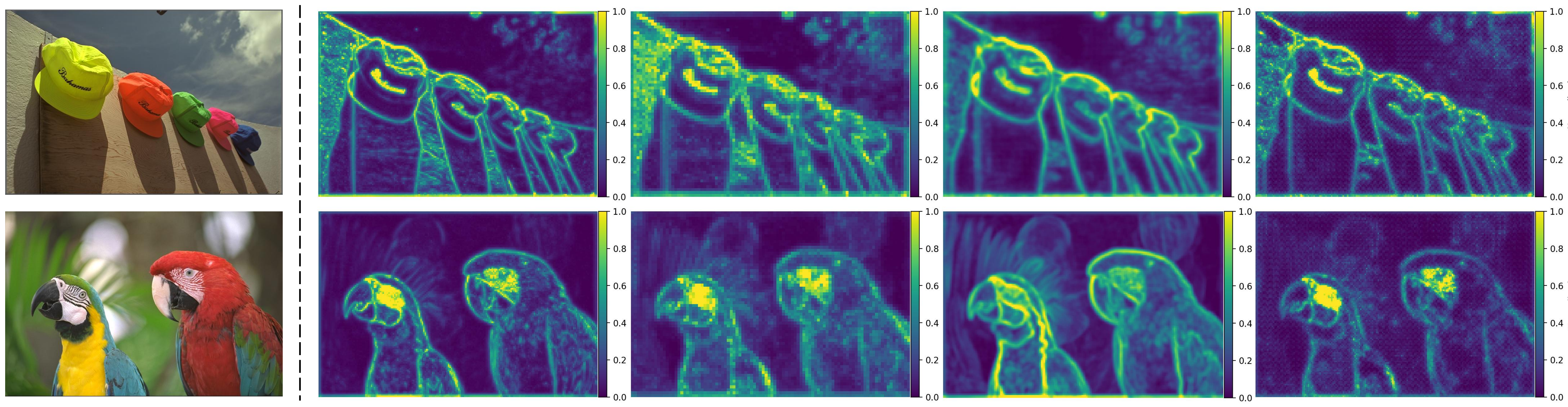}
\captionsetup{skip=1pt} 
\caption{\textbf{Spatial redundancy versus RMS-G scores.} Image023 from the Kodak dataset (left) and the RMS-G maps for the second- and third-stage graphs in GLIC (right). Brighter regions have higher RMS-G scores, indicating lower spatial redundancy and larger in-degrees.}
\vspace{-10pt}
\label{RMS-G}
\end{figure*}

\begin{figure*}[ht]
\centering
\includegraphics[width=1\textwidth]{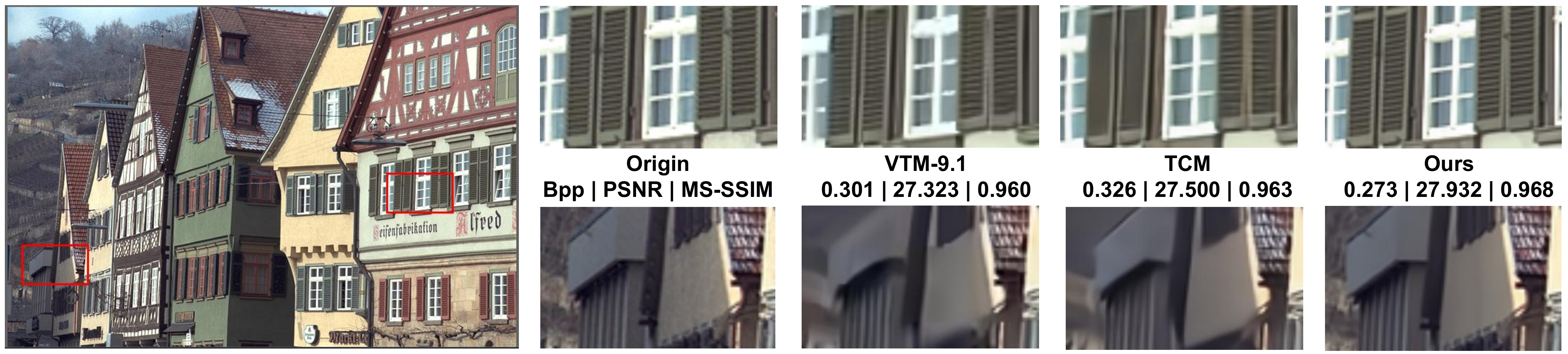}
\captionsetup{skip=1pt} 
\caption{Visual Comparisons with other methods. More results are in the supplementary.}
\vspace{-10pt}
\label{visual compare}
\end{figure*}

\vspace{-3pt}
\subsection{Computational Complexity}
\vspace{-3pt}
Beyond RD performance, we evaluate computational complexity on 2K-resolution images by reporting parameter count, forward FLOPs, decoding latency, and peak memory on an A100 GPU with an AMD EPYC 7742 64-core CPU, as summarized in Tab.~\ref{tab:bdrates_complexity}. GLIC attains a markedly better efficiency-latency trade-off than recent SOTA codecs. Compared with recent SOTA methods MambaIC \cite{zeng2025mambaic} (CVPR25), GLIC reduces parameter count, FLOPs, decoding latency and peak memory by 57.2\%, 55.4\%, 41.0\%, and 73.1\%, respectively. Relative to WeConvene, GLIC further reduces parameters by 36.3\%, FLOPs by 48.5\%, and decoding latency by 69.5\%. Compared with FTIC, it achieves substantially lower decoding latency with a similar parameter budget. These efficiency gains are obtained without sacrificing SOTA RD performance, indicating that the proposed GNN-based Feature Aggregation block effectively balances compression performance and computational cost.

\noindent\textbf{Computational overhead during graph construction.} We use $T=5$ iterations in Algorithm~\ref{alg:bisection} for graph construction. The updates are implemented with matrix operations and run in parallel over all nodes, so the overhead is small. The 5-iteration update accounts for about 5\% of the training time. At test time, the additional encoding and decoding latency on 2K images is below 0.02s (less than 5\% of the encoding/decoding times), confirming the efficiency of our bisection-based graph construction method.

\vspace{-3pt}
\subsection{Illustration of RMS-G Scores}
\vspace{-3pt}
We visualize the per-node RMS-G scores in Fig.~\ref{RMS-G}. Brighter colors indicate higher RMS-G values, corresponding to structurally complex, hard-to-compress regions that are assigned larger target degrees. In contrast, smooth background areas exhibit lower RMS-G values and thus receive fewer neighbors. Overall, the RMS-G map closely follows human perception by emphasizing high-frequency, texture-rich content where spatial redundancy is limited.

\vspace{-3pt}
\subsection{Ablation Studies}
\vspace{-3pt}

\begin{table}[t]
\centering
\setlength{\tabcolsep}{4pt}
\fontsize{9.1}{10}\selectfont
\renewcommand{\arraystretch}{1.15}
\caption{Ablations on dual-scale graph flexibility.}
\vspace{-8pt}
\begin{tabular}{l c c c}
\toprule
\textbf{Variant} & \textbf{Kodak} & \textbf{CLIC} & \textbf{Tecnick} \\
\midrule
w/o Global (Local only)   & -16.98 & -15.71 & -18.89 \\
w/o Local (Global only)  & -15.62 & -15.02 & -17.51 \\
Reverse (Local+Global)      & -18.77 & -18.21 & -20.79 \\
\midrule
\textbf{Global+Local (Ours)} & \textbf{-19.29} & \textbf{-18.71} & \textbf{-21.69} \\
\bottomrule
\end{tabular}
\vspace{-10pt}
\label{tab:abl_local_global}
\end{table}

\noindent\textbf{Dual-scale Graph Design.}
Tab.~\ref{tab:abl_local_global} studies the effect of the local and global graphs.
Using only the local graph or only the global graph leads to clear RD degradation on all three datasets.
The global-only variant is the weakest, showing that local aggregation is still necessary.
The ``Reverse'' setting, which flips the order of local and global aggregation, is better than the single-branch variants but remains worse than our design.
The full ``Global+Local'' configuration achieves the best BD-rates, especially on Tecnick, indicating that combining local details and long-range context in the proposed order is important for high-resolution images.

\noindent\textbf{Complexity-aware Scoring and Pooling.}
Tab.~\ref{tab:abl_scoring} evaluates different complexity scores and channel pooling strategies.
The row ``None / None'' disables complexity assessment, turning the model into a $k$NN-based GNN, which yields the weakest BD-rates.
Introducing any complexity score already brings consistent gains, and both Rescaling Residual and Sobel Gradient outperform local entropy.
Within the same score, channel pooling also matters: for Sobel Gradient, RMS pooling clearly improves over mean pooling on all three datasets.
The combination of Sobel Gradient and RMS pooling gives the strongest RD performance, and is therefore adopted as our default complexity measure.

 \begin{table}[t]
\centering
\setlength{\tabcolsep}{3pt}
\fontsize{9.1}{10}\selectfont
\renewcommand{\arraystretch}{1.15}
\caption{Ablations on complexity scoring and channel pooling.}
\vspace{-8pt}
\label{tab:abl_scoring}
\begin{tabular}{l l | c c c}
\toprule
\textbf{Scoring Strategy} & \textbf{Channel Pooling} & \textbf{Kodak} & \textbf{CLIC} & \textbf{Tecnick} \\
\midrule
None & None  & -16.97 & -16.21 & -18.21 \\
Local Entropy & RMS  & -17.05 & -17.01 & -18.97 \\
Rescaling Residual     & RMS & -17.67 & -17.03 & -19.68 \\
Rescaling Residual     & Mean & -18.23 & -17.82 & -20.39 \\
Sobel Gradient      & Mean  & -18.02 & -17.42 & -20.62 \\
\cellcolor{gray!20}\textbf{Sobel Gradient} & \cellcolor{gray!20}\textbf{RMS} & \cellcolor{gray!20}\textbf{-19.29} & \cellcolor{gray!20}\textbf{-18.71} & \cellcolor{gray!20}\textbf{-21.69} \\
\bottomrule
\end{tabular}
\vspace{-10pt}
\end{table}

\vspace{-3pt}
\subsection{Visual Comparisons}
\vspace{-3pt}
In Fig. \ref{visual compare}, we visually compare our GLIC model with the traditional codec VTM-9.1 \cite{bross2021overviewvvc} and LIC model TCM \cite{liu2023learned}. A detailed comparison of image patches reveals that our GLIC model better preserves high-frequency details and textures, effectively mitigating the severe detail loss and over-smoothing  observed in other methods. This improvement stems from our complexity-aware neighbor quota assignment mechanism, which adaptively allocates more connections to complex regions for enhanced feature aggregation.

More visualizations, ablations and discussions about limitations are provided in the supplementary.

\vspace{-3pt}
\section{Conclusion}
\vspace{-3pt}
We address the limitation that standard CNN kernels and window-based attention are inherently rigid. They impose fixed receptive fields and static connectivity that can couple uncorrelated features within the same Euclidean neighborhood and hinder adaptation to spatially varying redundancy. To overcome this, we propose GLIC, a content-adaptive LIC framework that builds dual-scale graphs and employs a complexity-aware scoring mechanism to enable flexible receptive fields and pixel-wise adaptive connectivity with linear complexity. Extensive experiments demonstrate state-of-the-art rate-distortion performance, validating the strong potential and effectiveness  of graph-based, content-adaptive redundancy modeling for image compression.

\section*{Acknowledgment}
This work was supported by the National Key Research and Development Program of China under Grant 2024YFF0509700,  National Natural Science Foundation of China (62471290,62431015,62331014,62501337) and the Fundamental Research Funds for the Central Universities.


{
    \small
    \bibliographystyle{ieeenat_fullname}
    \bibliography{main}

@String(CVPR= {IEEE Conf. Comput. Vis. Pattern Recog.})

@String(ICME = {Int. Conf. Multimedia and Expo})

@String(ICIP = {IEEE Int. Conf. Image Process.})

@String(AAAI = {AAAI})

@String(CVPR  = {CVPR})

@String(ICME  =	{ICME})

@String(ICIP  = {ICIP})

@inproceedings{
li2023frequency,
title={Frequency-Aware Transformer for Learned  Image Compression},
author={Han Li and Shaohui Li and Wenrui Dai and Chenglin Li and Junni Zou and Hongkai Xiong},
booktitle={The Twelfth International Conference on Learning Representations},
year={2024},
url={https://openreview.net/forum?id=HKGQDDTuvZ}
}

@inproceedings{liu2023learned,
  title={Learned image compression with mixed transformer-cnn architectures},
  author={Liu, Jinming and Sun, Heming and Katto, Jiro},
  booktitle={Proceedings of the IEEE/CVF conference on computer vision and pattern recognition},
  pages={14388--14397},
  year={2023}
}

@inproceedings{jiang2023mlic,
  title={Mlic: Multi-reference entropy model for learned image compression},
  author={Jiang, Wei and Yang, Jiayu and Zhai, Yongqi and Ning, Peirong and Gao, Feng and Wang, Ronggang},
  booktitle={Proceedings of the 31st ACM International Conference on Multimedia},
  pages={7618--7627},
  year={2023}
}

@inproceedings{
jiang2023mlic++,
title={{MLIC}++: Linear Complexity Multi-Reference Entropy Modeling for Learned Image Compression},
author={Wei Jiang and Ronggang Wang},
booktitle={ICML 2023 Workshop Neural Compression: From Information Theory to Applications},
year={2023},
url={https://openreview.net/forum?id=hxIpcSoz2t}
}

@inproceedings{zou2022devil,
  title={The devil is in the details: Window-based attention for image compression},
  author={Zou, Renjie and Song, Chunfeng and Zhang, Zhaoxiang},
  booktitle={Proceedings of the IEEE/CVF conference on computer vision and pattern recognition},
  pages={17492--17501},
  year={2022}
}

@inproceedings{zhu2022transformer,
  title={Transformer-based transform coding},
  author={Zhu, Yinhao and Yang, Yang and Cohen, Taco},
  booktitle={International Conference on Learning Representations},
  year={2022}
}

@inproceedings{
balle2018variational,
title={Variational image compression with a scale hyperprior},
author={Johannes Ballé and David Minnen and Saurabh Singh and Sung Jin Hwang and Nick Johnston},
booktitle={International Conference on Learning Representations},
year={2018},
url={https://openreview.net/forum?id=rkcQFMZRb},
}

@article{minnen2018joint,
  title={Joint autoregressive and hierarchical priors for learned image compression},
  author={Minnen, David and Ball{\'e}, Johannes and Toderici, George D},
  journal={Advances in neural information processing systems},
  volume={31},
  year={2018}
}

@inproceedings{minnen2020channel,
  title={Channel-wise autoregressive entropy models for learned image compression},
  author={Minnen, David and Singh, Saurabh},
  booktitle={2020 IEEE International Conference on Image Processing (ICIP)},
  pages={3339--3343},
  year={2020},
  organization={IEEE}
}

@inproceedings{he2022elic,
  title={Elic: Efficient learned image compression with unevenly grouped space-channel contextual adaptive coding},
  author={He, Dailan and Yang, Ziming and Peng, Weikun and Ma, Rui and Qin, Hongwei and Wang, Yan},
  booktitle={Proceedings of the IEEE/CVF Conference on Computer Vision and Pattern Recognition},
  pages={5718--5727},
  year={2022}
}

@inproceedings{cheng2020learned,
  title={Learned image compression with discretized gaussian mixture likelihoods and attention modules},
  author={Cheng, Zhengxue and Sun, Heming and Takeuchi, Masaru and Katto, Jiro},
  booktitle={Proceedings of the IEEE/CVF conference on computer vision and pattern recognition},
  pages={7939--7948},
  year={2020}
}

@InProceedings{qian2022entroformer,
    author    = {Qian, Yichen and Lin, Ming and Sun, Xiuyu and Tan, Zhiyu and Jin, Rong},
    title     = {Entroformer: A Transformer-based Entropy Model for Learned Image Compression},
    booktitle = {International Conference on Learning Representations},
    month     = {May},
    year      = {2022},
}

@inproceedings{xie2021enhanced,
  title={Enhanced invertible encoding for learned image compression},
  author={Xie, Yueqi and Cheng, Ka Leong and Chen, Qifeng},
  booktitle={Proceedings of the 29th ACM international conference on multimedia},
  pages={162--170},
  year={2021}
}

@inproceedings{chen2022two,
  title={Two-stage octave residual network for end-to-end image compression},
  author={Chen, Fangdong and Xu, Yumeng and Wang, Li},
  booktitle={Proceedings of the AAAI Conference on Artificial Intelligence},
  volume={36},
  number={4},
  pages={3922--3929},
  year={2022}
}

@inproceedings{
balle2016end,
title={End-to-end Optimized Image Compression},
author={Johannes Ball{\'e} and Valero Laparra and Eero P. Simoncelli},
booktitle={International Conference on Learning Representations},
year={2017},
url={https://openreview.net/forum?id=rJxdQ3jeg}
}

@inproceedings{he2021checkerboard,
  title={Checkerboard context model for efficient learned image compression},
  author={He, Dailan and Zheng, Yaoyan and Sun, Baocheng and Wang, Yan and Qin, Hongwei},
  booktitle={Proceedings of the IEEE/CVF Conference on Computer Vision and Pattern Recognition},
  pages={14771--14780},
  year={2021}
}

@inproceedings{li2022hybrid,
  title={Hybrid spatial-temporal entropy modelling for neural video compression},
  author={Li, Jiahao and Li, Bin and Lu, Yan},
  booktitle={Proceedings of the 30th ACM International Conference on Multimedia},
  pages={1503--1511},
  year={2022}
}

@inproceedings{li2023neural,
  title={Neural video compression with diverse contexts},
  author={Li, Jiahao and Li, Bin and Lu, Yan},
  booktitle={Proceedings of the IEEE/CVF Conference on Computer Vision and Pattern Recognition},
  pages={22616--22626},
  year={2023}
}

@article{luo2016understanding,
  title={Understanding the effective receptive field in deep convolutional neural networks},
  author={Luo, Wenjie and Li, Yujia and Urtasun, Raquel and Zemel, Richard},
  journal={Advances in neural information processing systems},
  volume={29},
  year={2016}
}

@article{liu2020unified,
  title={A unified end-to-end framework for efficient deep image compression},
  author={Liu, Jiaheng and Lu, Guo and Hu, Zhihao and Xu, Dong},
  journal={arXiv preprint arXiv:2002.03370},
  year={2020}
}

@inproceedings{bjontegaard2001,
  author    = {Gisle Bjontegaard},
  title     = {Calculation of average PSNR differences between RD-curves},
  booktitle = {VCEG-M33},
  year      = {2001}
}

@misc{kodak1993,
  author       = {Eastman Kodak},
  title        = {Kodak Lossless True Color Image Suite (PhotoCD PCD0992)},
  year         = {1993},
  note         = {Available from http://r0k.us/graphics/kodak/}
}

@inproceedings{asuni2014testimages,
  title={TESTIMAGES: a Large-scale Archive for Testing Visual Devices and Basic Image Processing Algorithms.},
  author={Asuni, Nicola and Giachetti, Andrea and others},
  booktitle={STAG},
  pages={63--70},
  year={2014}
}

@article{sobel19683x3,
  title={A 3x3 isotropic gradient operator for image processing},
  author={Sobel, Irwin and Feldman, Gary and others},
  journal={a talk at the Stanford Artificial Project in},
  volume={1968},
  pages={271--272},
  year={1968}
}

@article{jiang2025mlicv2,
  title={MLICv2: Enhanced Multi-Reference Entropy Modeling for Learned Image Compression},
  author={Jiang, Wei and Zhai, Yongqi and Yang, Jiayu and Gao, Feng and Wang, Ronggang},
  journal={arXiv preprint arXiv:2504.19119},
  year={2025}
}

@article{qin2024mambavc,
  title={MambaVC: Learned Visual Compression with Selective State Spaces},
  author={Qin, Shiyu and Wang, Jinpeng and Zhou, Yimin and Chen, Bin and Luo, Tianci and An, Baoyi and Dai, Tao and Xia, Shutao and Wang, Yaowei},
  journal={arXiv preprint arXiv:2405.15413},
  year={2024}
}

@article{kingma2014adam,
  title={Adam: A method for stochastic optimization},
  author={Kingma, Diederik P},
  journal={International Conference on Learning Representations},
  year={2015}
}

@article{bross2021overviewvvc,
  title={Overview of the versatile video coding (VVC) standard and its applications},
  author={Bross, Benjamin and Wang, Ye-Kui and Ye, Yan and Liu, Shan and Chen, Jianle and Sullivan, Gary J and Ohm, Jens-Rainer},
  journal={IEEE Transactions on Circuits and Systems for Video Technology},
  volume={31},
  number={10},
  pages={3736--3764},
  year={2021},
  publisher={IEEE}
}

@inproceedings{zhang2023neural,
  title={Neural Rate Control for Learned Video Compression},
  author={Zhang, Yiwei and Lu, Guo and Chen, Yunuo and Wang, Shen and Shi, Yibo and Wang, Jing and Song, Li},
  booktitle={The Twelfth International Conference on Learning Representations},
  year={2023}
}

@misc{CLIC2020,
  title = {Workshop and Challenge on Learned Image Compression (CLIC2020)},
  author = {George Toderici and Wenzhe Shi and Radu Timofte and Lucas Theis and Johannes Balle and Eirikur Agustsson and Nick Johnston and Fabian Mentzer},
  url = {http://www.compression.cc},
  year={2020},
  organization={CVPR}
}

@article{llic,
  title={Llic: Large receptive field transform coding with adaptive weights for learned image compression},
  author={Jiang, Wei and Ning, Peirong and Yang, Jiayu and Zhai, Yongqi and Gao, Feng and Wang, Ronggang},
  journal={IEEE Transactions on Multimedia},
  volume={26},
  pages={10937--10951},
  year={2024},
  publisher={IEEE}
}

@inproceedings{zafari2023frequency-hilo,
  title={Frequency disentangled features in neural image compression},
  author={Zafari, Ali and Khoshkhahtinat, Atefeh and Mehta, Piyush and Saadabadi, Mohammad Saeed Ebrahimi and Akyash, Mohammad and Nasrabadi, Nasser M},
  booktitle={2023 IEEE International Conference on Image Processing (ICIP)},
  pages={2815--2819},
  year={2023},
  organization={IEEE}
}

@inproceedings{
mentzer2022vct,
title={{VCT}: A Video Compression Transformer},
author={Fabian Mentzer and George Toderici and David Minnen and Sergi Caelles and Sung Jin Hwang and Mario Lucic and Eirikur Agustsson},
booktitle={Advances in Neural Information Processing Systems},
editor={Alice H. Oh and Alekh Agarwal and Danielle Belgrave and Kyunghyun Cho},
year={2022},
url={https://openreview.net/forum?id=lme1MKnSMb}
}

@article{ma2019iwave,
  title={iWave: CNN-based wavelet-like transform for image compression},
  author={Ma, Haichuan and Liu, Dong and Xiong, Ruiqin and Wu, Feng},
  journal={IEEE Transactions on Multimedia},
  volume={22},
  number={7},
  pages={1667--1679},
  year={2019},
  publisher={IEEE}
}

@inproceedings{lu2022transformer,
  title={Transformer-based Image Compression},
  author={Lu, Ming and Guo, Peiyao and Shi, Huiqing and Cao, Chuntong and Ma, Zhan},
  booktitle={2022 Data Compression Conference (DCC)},
  pages={469--469},
  year={2022},
  organization={IEEE}
}

@inproceedings{gao2021neural,
  title={Neural image compression via attentional multi-scale back projection and frequency decomposition},
  author={Gao, Ge and You, Pei and Pan, Rong and Han, Shunyuan and Zhang, Yuanyuan and Dai, Yuchao and Lee, Hojae},
  booktitle={Proceedings of the IEEE/CVF International Conference on Computer Vision},
  pages={14677--14686},
  year={2021}
}

@article{fu2023asymmetric,
  title={Asymmetric learned image compression with multi-scale residual block, importance scaling, and post-quantization filtering},
  author={Fu, Haisheng and Liang, Feng and Liang, Jie and Li, Binglin and Zhang, Guohe and Han, Jingning},
  journal={IEEE Transactions on Circuits and Systems for Video Technology},
  volume={33},
  number={8},
  pages={4309--4321},
  year={2023},
  publisher={IEEE}
}

@article{begaint2020compressai,
  title={CompressAI: a PyTorch library and evaluation platform for end-to-end compression research},
  author={B{\'e}gaint, Jean and Racap{\'e}, Fabien and Feltman, Simon and Pushparaja, Akshay},
  journal={arXiv e-prints},
  pages={arXiv--2011},
  year={2020}
}

@inproceedings{landrieu2018large,
  title={Large-scale point cloud semantic segmentation with superpoint graphs},
  author={Landrieu, Loic and Simonovsky, Martin},
  booktitle={Proceedings of the IEEE conference on computer vision and pattern recognition},
  pages={4558--4567},
  year={2018}
}

@inproceedings{yan2018spatial,
  title={Spatial temporal graph convolutional networks for skeleton-based action recognition},
  author={Yan, Sijie and Xiong, Yuanjun and Lin, Dahua},
  booktitle={Proceedings of the AAAI conference on artificial intelligence},
  volume={32},
  number={1},
  year={2018}
}

@inproceedings{
ma2023image,
title={Image as Set of Points},
author={Xu Ma and Yuqian Zhou and Huan Wang and Can Qin and Bin Sun and Chang Liu and Yun Fu},
booktitle={The Eleventh International Conference on Learning Representations },
year={2023},
url={https://openreview.net/forum?id=awnvqZja69}
}

@article{liu2022dual,
  title={Dual learning-based graph neural network for remote sensing image super-resolution},
  author={Liu, Ziyu and Feng, Ruyi and Wang, Lizhe and Han, Wei and Zeng, Tieyong},
  journal={IEEE Transactions on Geoscience and Remote Sensing},
  volume={60},
  pages={1--14},
  year={2022},
  publisher={IEEE}
}

@article{zhou2020cross,
  title={Cross-scale internal graph neural network for image super-resolution},
  author={Zhou, Shangchen and Zhang, Jiawei and Zuo, Wangmeng and Loy, Chen Change},
  journal={Advances in neural information processing systems},
  volume={33},
  pages={3499--3509},
  year={2020}
}

@inproceedings{li2021cross,
  title={Cross-patch graph convolutional network for image denoising},
  author={Li, Yao and Fu, Xueyang and Zha, Zheng-Jun},
  booktitle={Proceedings of the IEEE/CVF International Conference on Computer Vision},
  pages={4651--4660},
  year={2021}
}

@inproceedings{mou2021graph,
  title={Graph attention neural network for image restoration},
  author={Mou, Chong and Zhang, Jian},
  booktitle={2021 IEEE International Conference on Multimedia and Expo (ICME)},
  pages={1--6},
  year={2021},
  organization={IEEE}
}

@inproceedings{tian2024image,
  title={Image Processing GNN: Breaking Rigidity in Super-Resolution},
  author={Tian, Yuchuan and Chen, Hanting and Xu, Chao and Wang, Yunhe},
  booktitle={Proceedings of the IEEE/CVF Conference on Computer Vision and Pattern Recognition},
  pages={24108--24117},
  year={2024}
}

@article{wu2025conditional,
  title={Conditional Latent Coding with Learnable Synthesized Reference for Deep Image Compression},
  author={Wu, Siqi and Chen, Yinda and Liu, Dong and He, Zhihai},
  journal={arXiv preprint arXiv:2502.09971},
  year={2025}
}

@article{han2022vision,
  title={Vision gnn: An image is worth graph of nodes},
  author={Han, Kai and Wang, Yunhe and Guo, Jianyuan and Tang, Yehui and Wu, Enhua},
  journal={Advances in neural information processing systems},
  volume={35},
  pages={8291--8303},
  year={2022}
}

@inproceedings{han2023vision,
  title={Vision hgnn: An image is more than a graph of nodes},
  author={Han, Yan and Wang, Peihao and Kundu, Souvik and Ding, Ying and Wang, Zhangyang},
  booktitle={Proceedings of the IEEE/CVF International Conference on Computer Vision},
  pages={19878--19888},
  year={2023}
}

@inproceedings{gori2005new,
  title={A new model for learning in graph domains},
  author={Gori, Marco and Monfardini, Gabriele and Scarselli, Franco},
  booktitle={Proceedings. 2005 IEEE international joint conference on neural networks, 2005.},
  volume={2},
  pages={729--734},
  year={2005},
  organization={IEEE}
}

@article{micheli2009neural,
  title={Neural network for graphs: A contextual constructive approach},
  author={Micheli, Alessio},
  journal={IEEE Transactions on Neural Networks},
  volume={20},
  number={3},
  pages={498--511},
  year={2009},
  publisher={IEEE}
}

@inproceedings{xu2017scene,
  title={Scene graph generation by iterative message passing},
  author={Xu, Danfei and Zhu, Yuke and Choy, Christopher B and Fei-Fei, Li},
  booktitle={Proceedings of the IEEE conference on computer vision and pattern recognition},
  pages={5410--5419},
  year={2017}
}

@inproceedings{jing2022learning,
  title={Learning graph neural networks for image style transfer},
  author={Jing, Yongcheng and Mao, Yining and Yang, Yiding and Zhan, Yibing and Song, Mingli and Wang, Xinchao and Tao, Dacheng},
  booktitle={European Conference on Computer Vision},
  pages={111--128},
  year={2022},
  organization={Springer}
}

@inproceedings{yang2020distilling,
  title={Distilling knowledge from graph convolutional networks},
  author={Yang, Yiding and Qiu, Jiayan and Song, Mingli and Tao, Dacheng and Wang, Xinchao},
  booktitle={Proceedings of the IEEE/CVF conference on computer vision and pattern recognition},
  pages={7074--7083},
  year={2020}
}

@misc{chen2025s2cformerreorientinglearnedimage,
      title={S2CFormer: Reorienting Learned Image Compression from Spatial Interaction to Channel Aggregation}, 
      author={Yunuo Chen and Qian Li and Bing He and Donghui Feng and Ronghua Wu and Qi Wang and Li Song and Guo Lu and Wenjun Zhang},
      year={2025},
      eprint={2502.00700},
      archivePrefix={arXiv},
      primaryClass={cs.CV},
      url={https://arxiv.org/abs/2502.00700}, 
}

@inproceedings{
han2024causal,
title={Causal Context Adjustment Loss for Learned Image Compression},
author={Minghao Han and Shiyin Jiang and Shengxi Li and Xin Deng and Mai Xu and Ce Zhu and Shuhang Gu},
booktitle={The Thirty-eighth Annual Conference on Neural Information Processing Systems},
year={2024},
url={https://openreview.net/forum?id=AYntCZvoLI}
}

@article{DBLP:journals/corr/YuK15,
  title={Multi-scale context aggregation by dilated convolutions},
  author={Yu, Fisher and Koltun, Vladlen},
  journal={arXiv preprint arXiv:1511.07122},
  year={2015}
}

@misc{hassani2023dilatedneighborhoodattentiontransformer,
      title={Dilated Neighborhood Attention Transformer}, 
      author={Ali Hassani and Humphrey Shi},
      year={2023},
      eprint={2209.15001},
      archivePrefix={arXiv},
      primaryClass={cs.CV},
      url={https://arxiv.org/abs/2209.15001}, 
}

@inproceedings{simonovsky2017dynamic,
  title={Dynamic edge-conditioned filters in convolutional neural networks on graphs},
  author={Simonovsky, Martin and Komodakis, Nikos},
  booktitle={Proceedings of the IEEE conference on computer vision and pattern recognition},
  pages={3693--3702},
  year={2017}
}

@inproceedings{
li2025on,
title={On Disentangled Training for Nonlinear Transform in Learned Image Compression},
author={Han Li and Shaohui Li and Wenrui Dai and Maida Cao and Nuowen Kan and Chenglin Li and Junni Zou and Hongkai Xiong},
booktitle={The Thirteenth International Conference on Learning Representations},
year={2025},
url={https://openreview.net/forum?id=U67J0QNtzo}
}

@article{zeng2025mambaic,
  title={MambaIC: State Space Models for High-Performance Learned Image Compression},
  author={Zeng, Fanhu and Tang, Hao and Shao, Yihua and Chen, Siyu and Shao, Ling and Wang, Yan},
  journal={arXiv preprint arXiv:2503.12461},
  year={2025}
}

@article{dcnic,
  title={Deep image compression based on multi-scale deformable convolution},
  author={Li, Daowen and Li, Yingming and Sun, Heming and Yu, Lu},
  journal={Journal of Visual Communication and Image Representation},
  volume={87},
  pages={103573},
  year={2022},
  publisher={Elsevier}
}

@article{fu2024fast,
  title={Fast and high-performance learned image compression with improved checkerboard context model, deformable residual module, and knowledge distillation},
  author={Fu, Haisheng and Liang, Feng and Liang, Jie and Wang, Yongqiang and Fang, Zhenman and Zhang, Guohe and Han, Jingning},
  journal={IEEE Transactions on Image Processing},
  year={2024},
  publisher={IEEE}
}

@article{dcnvc,
  title={Learned video compression via heterogeneous deformable compensation network},
  author={Wang, Huairui and Chen, Zhenzhong and Chen, Chang Wen},
  journal={IEEE Transactions on Multimedia},
  volume={26},
  pages={1855--1866},
  year={2023},
  publisher={IEEE}
}

@article{tang2022jointgat,
  title={Joint graph attention and asymmetric convolutional neural network for deep image compression},
  author={Tang, Zhisen and Wang, Hanli and Yi, Xiaokai and Zhang, Yun and Kwong, Sam and Kuo, C-C Jay},
  journal={IEEE Transactions on Circuits and Systems for Video Technology},
  volume={33},
  number={1},
  pages={421--433},
  year={2022},
  publisher={IEEE}
}

@inproceedings{spadaro2024gabic,
  title={Gabic: Graph-based attention block for image compression},
  author={Spadaro, Gabriele and Presta, Alberto and Tartaglione, Enzo and Giraldo, Jhony H and Grangetto, Marco and Fiandrotti, Attilio},
  booktitle={2024 IEEE International Conference on Image Processing (ICIP)},
  pages={1802--1808},
  year={2024},
  organization={IEEE}
}

@inproceedings{gcnic,
  title={Graph-convolution network for image compression},
  author={Yang, Chunhui and Ma, Yi and Yang, Jiayu and Liu, Shiyi and Wang, Ronggang},
  booktitle={2021 IEEE International Conference on Image Processing (ICIP)},
  pages={2094--2098},
  year={2021},
  organization={IEEE}
}

@inproceedings{akyazi2019learning-anotherwave1,
  title={Learning-based image compression using convolutional autoencoder and wavelet decomposition},
  author={Akyazi, Pinar and Ebrahimi, Touradj},
  booktitle={IEEE Conference on Computer Vision and Pattern Recognition Workshops},
  year={2019}
}

@article{fu2023learned-octaveFU1,
  title={Learned image compression with generalized octave convolution and cross-resolution parameter estimation},
  author={Fu, Haisheng and Liang, Feng},
  journal={Signal Processing},
  volume={202},
  pages={108778},
  year={2023},
  publisher={Elsevier}
}

@inproceedings{fu2024weconvene,
  title={WeConvene: Learned Image Compression with Wavelet-Domain Convolution and Entropy Model},
  author={Fu, Haisheng and Liang, Jie and Fang, Zhenman and Han, Jingning and Liang, Feng and Zhang, Guohe},
  booktitle={European Conference on Computer Vision},
  pages={37--53},
  year={2024},
  organization={Springer}
}

@inproceedings{zhang2024another,
  title={Another way to the top: Exploit contextual clustering in learned image coding},
  author={Zhang, Yichi and Duan, Zhihao and Lu, Ming and Ding, Dandan and Zhu, Fengqing and Ma, Zhan},
  booktitle={Proceedings of the AAAI Conference on Artificial Intelligence},
  volume={38},
  number={8},
  pages={9377--9386},
  year={2024}
}

@article{iccv25HPCM,
  title={Learned image compression with hierarchical progressive context modeling},
  author={Li, Yuqi and Zhang, Haotian and Li, Li and Liu, Dong},
  journal={arXiv preprint arXiv:2507.19125},
  year={2025}
}

@inproceedings{cvpr2025diction_dcae,
  title={Learned Image Compression with Dictionary-based Entropy Model},
  author={Lu, Jingbo and Zhang, Leheng and Zhou, Xingyu and Li, Mu and Li, Wen and Gu, Shuhang},
  booktitle={Proceedings of the Computer Vision and Pattern Recognition Conference},
  pages={12850--12859},
  year={2025}
}

@inproceedings{lalic,
  title={Linear Attention Modeling for Learned Image Compression},
  author={Feng, Donghui and Cheng, Zhengxue and Wang, Shen and Wu, Ronghua and Hu, Hongwei and Lu, Guo and Song, Li},
  booktitle={Proceedings of the Computer Vision and Pattern Recognition Conference},
  pages={7623--7632},
  year={2025}
}

@article{tian2025smc++,
  title={Smc++: Masked learning of unsupervised video semantic compression},
  author={Tian, Yuan and Ling, Xiaoyue and Geng, Cong and Hu, Qiang and Lu, Guo and Zhai, Guangtao},
  journal={IEEE Transactions on Pattern Analysis and Machine Intelligence},
  year={2025},
  publisher={IEEE}
}

@article{tian2024coding,
  title={A coding framework and benchmark towards low-bitrate video understanding},
  author={Tian, Yuan and Lu, Guo and Yan, Yichao and Zhai, Guangtao and Chen, Li and Gao, Zhiyong},
  journal={IEEE Transactions on Pattern Analysis and Machine Intelligence},
  volume={46},
  number={8},
  pages={5852--5872},
  year={2024},
  publisher={IEEE}
}

@article{du2025largeLLM,
  title={Large language model for lossless image compression with visual prompts},
  author={Du, Junhao and Zhou, Chuqin and Cao, Ning and Chen, Gang and Chen, Yunuo and Cheng, Zhengxue and Song, Li and Lu, Guo and Zhang, Wenjun},
  journal={arXiv preprint arXiv:2502.16163},
  year={2025}
}

@inproceedings{chen2025knowledgeKDIC,
  title={Knowledge Distillation for Learned Image Compression},
  author={Chen, Yunuo and Lyu, Zezheng and He, Bing and Cao, Ning and Chen, Gang and Lu, Guo and Zhang, Wenjun},
  booktitle={Proceedings of the IEEE/CVF International Conference on Computer Vision},
  pages={4996--5006},
  year={2025}
}

@inproceedings{li2025differentiableVMAF,
  title={Differentiable VMAF: A trainable metric for optimizing video compression codec},
  author={Li, Jiangchuan and Zhou, Chuqin and Chen, Yunuo and Lu, Guo},
  booktitle={2025 IEEE International Symposium on Circuits and Systems (ISCAS)},
  pages={1--5},
  year={2025},
  organization={IEEE}
}

@article{liang2025structureGS,
  title={Structure-Guided Allocation of 2D Gaussians for Image Representation and Compression},
  author={Liang, Huanxiong and Chen, Yunuo and Pan, Yicheng and Wang, Sixian and Dai, Jincheng and Lu, Guo and Zhang, Wenjun},
  journal={arXiv preprint arXiv:2512.24018},
  year={2025}
}

@article{ling2026free,
  title={Free-GVC: Towards Training-Free Extreme Generative Video Compression with Temporal Coherence},
  author={Ling, Xiaoyue and Zhou, Chuqin and Li, Chunyi and Chen, Yunuo and Tian, Yuan and Lu, Guo and Zhang, Wenjun},
  journal={arXiv preprint arXiv:2602.09868},
  year={2026}
}

@inproceedings{
chen2026contentaware,
title={Content-Aware Mamba for Learned Image Compression},
author={Yunuo Chen and Zezheng Lyu and Bing He and Hongwei Hu and Qi Wang and Yuan Tian and Li Song and Wenjun Zhang and Guo Lu},
booktitle={The Fourteenth International Conference on Learning Representations},
year={2026},
url={https://openreview.net/forum?id=WwDNiisZQm}
}

@article{zhou2026dual,
  title={Dual-Representation Image Compression at Ultra-Low Bitrates via Explicit Semantics and Implicit Textures},
  author={Zhou, Chuqin and Ling, Xiaoyue and Chen, Yunuo and Dai, Jincheng and Lu, Guo and Zhang, Wenjun},
  journal={arXiv preprint arXiv:2602.05213},
  year={2026}
}

@inproceedings{lu2025vcip,
  title={VCIP 2025 Ultra Low-Bitrate Video Compression Challenge},
  author={Lu, Guo and Wang, Jing and Chen, Yunuo and Zhou, Chuqin and Shi, Yibo and Lin, Kai},
  booktitle={2025 International Conference on Visual Communications and Image Processing (VCIP)},
  pages={1--3},
  year={2025},
  organization={IEEE}
}

@article{chen2026nextframe,
  title         = {Next-Frame Decoding for Ultra-Low-Bitrate Image Compression with Video Diffusion Priors},
  author        = {Chen, Yunuo and Zhou, Chuqin and Li, Jiangchuan and Ling, Xiaoyue and He, Bing and Dai, Jincheng and Song, Li and Lu, Guo},
  journal       = {arXiv preprint arXiv:2603.15129},
  year          = {2026},
  eprint        = {2603.15129},
  archivePrefix = {arXiv},
  primaryClass  = {cs.CV},
  doi           = {10.48550/arXiv.2603.15129},
  url           = {https://arxiv.org/abs/2603.15129}
}
}


\end{document}